\documentclass[runningheads]{llncs}

\usepackage{eccv}

\usepackage{eccvabbrv}

\usepackage{graphicx}
\usepackage{booktabs}
\usepackage{multirow}
\usepackage{pifont}
\usepackage{tabularray}
\usepackage{caption}

\usepackage[accsupp]{axessibility}  %

\usepackage[pagebackref,breaklinks,colorlinks]{hyperref}

\usepackage{orcidlink}
\newcommand\mc[1]{\multicolumn{1}{c}{#1}} %

\newcommand{\projectname}[0]{LayeredFlow}

\begin{document}

\title{\projectname{}: A Real-World Benchmark for Non-Lambertian Multi-Layer Optical Flow} 

\titlerunning{\projectname{}}

\author{Hongyu Wen \orcidlink{0009-0002-6526-6283} \and
Erich Liang \orcidlink{0000-0001-7414-0194}  \and
Jia Deng \orcidlink{0000-0001-9594-4554}}

\authorrunning{H. Wen et al.}

\institute{Department of Computer Science, Princeton University\\
\email{\{hongyu.wen, erliang, jiadeng\}@princeton.edu}}

\maketitle

\begin{abstract}
Achieving 3D understanding of non-Lambertian objects is an important task with many useful applications, but most existing algorithms struggle to deal with such objects. One major obstacle towards progress in this field is the lack of holistic non-Lambertian benchmarks---most benchmarks have low scene and object diversity, and none provide multi-layer 3D annotations for objects occluded by transparent surfaces. In this paper, we introduce \projectname{}, a real world benchmark containing multi-layer ground truth annotation for optical flow of non-Lambertian objects. Compared to previous benchmarks, our benchmark exhibits greater scene and object diversity, with 150k high quality optical flow and stereo pairs taken over 185 indoor and outdoor scenes and 360 unique objects. Using \projectname{} as evaluation data, we propose a new task called multi-layer optical flow. To provide training data for this task, we introduce a large-scale densely-annotated synthetic dataset containing 60k images within 30 scenes tailored for non-Lambertian objects. Training on our synthetic dataset enables model to predict multi-layer optical flow, while fine-tuning existing optical flow methods on the dataset notably boosts their performance on non-Lambertian objects without compromising the performance on diffuse objects. Data is available at \href{https://layeredflow.cs.princeton.edu}{https://layeredflow.cs.princeton.edu}.
\end{abstract}

\section{Introduction}
\label{sec:intro}

Achieving 3D understanding of non-Lambertian objects is an important task because they appear in many real world applications.
In autonomous navigation, correct 3D geometry of glass walls and reflective road patches is crucial for path planning. In robotics, accurate depth information for plastic and metal materials is necessary for precise and dexterous manipulation of common household items. 

However, many techniques which perform well on diffuse surfaces struggle to capture accurate information about non-Lambertian surfaces. Conventional depth measurement methods, like lighting \cite{middleburry} and time-of-flight techniques \cite{kitti1, hd1k, eth3d}, are highly sensitive to the reflective properties of surfaces and cannot generate reliable 3D information for non-Lambertian objects. Similarly, data-driven algorithms for problems like optical flow \cite{raft, flowformer, croco_v2} and stereo matching \cite{raft-stereo, DLNR, hitnet} perform well for diffuse objects, but often catastrophically fail for non-Lambertian objects. This is because most image datasets used for training \cite{infinigen, mayer2016large} or evaluation \cite{sintel, spring, eth3d} contain significantly more Lambertian objects than non-Lambertian objects, which downplays the importance of non-Lambertian objects within the benchmark evaluation.

\begin{figure*}[t]
    \centering
    \includegraphics[width=0.8\linewidth]{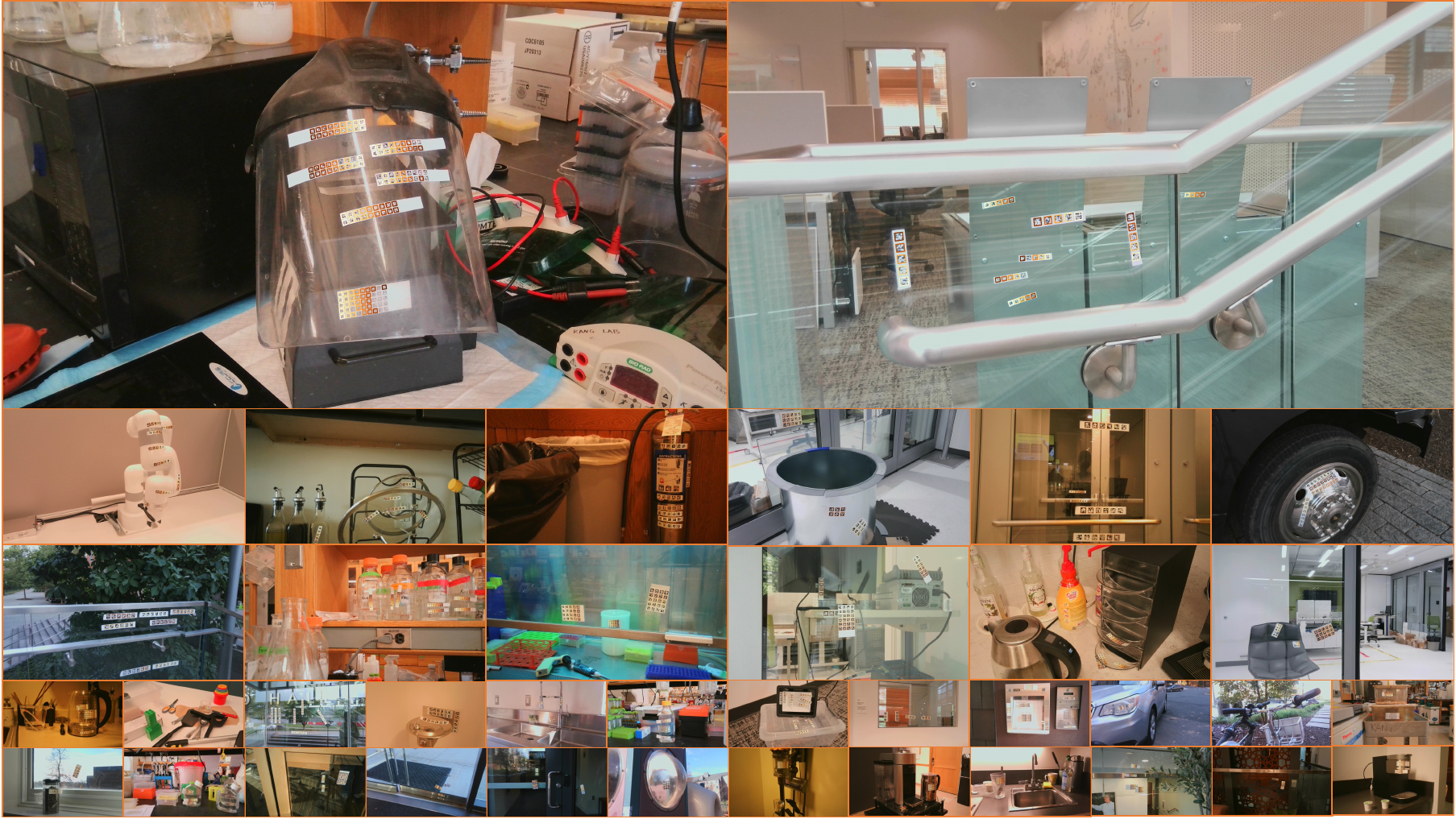}
    \caption{Gallery of our non-Lambertian real world benchmark. Our benchmark encompasses 185 indoor and outdoor scenes and 360 different objects with 2000 images. By using a stereo camera and carefully attaching and removing AprilTags, we acquire accurate multi-layer optical flow and stereo measurements.}
    \label{fig:benchmark_gallery}
    \vspace{-5mm}
\end{figure*}

In addition, existing real world benchmarks designed directly for non-Lamber-tian objects suffer from data diversity issues. Most benchmark scenes are confined to a limited number of indoor environments. Additionally, the non-Lambertian objects are typically small, tabletop objects. This is because in order to obtain accurate 3D ground truth data for these objects, some works \cite{clearpose, transcg, phocal, seeingglass} utilize pre-scanned 3D models of the objects they use, while other works \cite{booster, glasswall} physically paint their objects with Lambertian paint to aid correspondence detection. These benchmark design choices limit the diversity of potential objects and make the data collection process hard to scale.

More importantly, existing non-Lambertian benchmarks do not provide multi-layer 3D geometry data when transparent surfaces are present.
However, when imaging transparent objects, an individual pixel can now capture information about multiple 3D points in the scene: one point on a transparent surface \emph{as well as points on occluded objects behind it}. In these settings, humans are often able to infer 3D information at multiple layers of depth. If we wish to build algorithms with a similar level of 3D scene comprehension, it is imperative to include multi-layer 3D data in our real world benchmarks.

\begin{figure*}[t]
    \centering
    \includegraphics[width=0.8\linewidth]{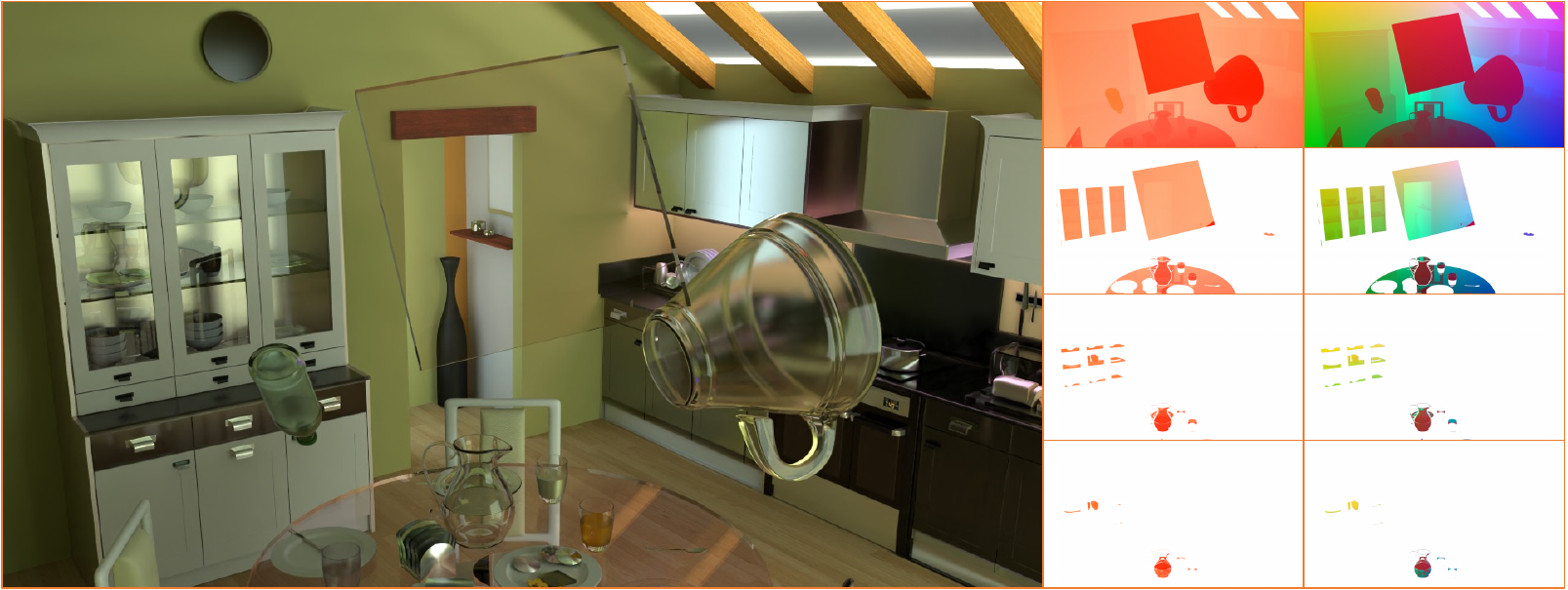}
    \caption{Showcase of our synthetic dataset with ground truth annotations. Left: a sample synthetic image. Right: Multi-layer optical flow and 3D positions in world coordinates.}
    \label{fig:synthetic_gt}
\end{figure*}

In this work, we introduce \projectname{}, our real world benchmark of non-Lambertian objects with ground truth annotation of multi-layer 3D geometry. We use AprilTag \cite{apriltags}, a visual fiducial system, as well as a stereo camera system to triangulate tagged 3D points within a scene. This allows for the capture of multi-layer ground truth geometry, as tags positioned behind transparent objects still remain detectable by cameras. We capture each scene with our stereo camera system before and after a rearrangement of objects in the scene. Using this method, we capture 2k images of 360 distinct objects in 185 indoor and outdoor scenes, ranging from household to laboratory settings. Using the tagged version of images, we generate 150k high quality ground truth optical flow and stereo pair annotations. Our benchmark is significantly more diverse than any existing non-Lambertian objects datasets and is the only one that has real world optical flow annotations. A gallery of our benchmark is showcased in \cref{fig:benchmark_gallery}.

Using \projectname{} as evaluation data, we propose a new task called multi-layer optical flow, which requires flow prediction for all visible surfaces even when they are behind transparent objects. 
To provide training data for this task, we introduce a large-scale synthetic dataset that contains 60k images on 30 high-quality artist-made indoor scenes, with multi-layer optical flow and 3D position ground truth. See \cref{fig:synthetic_gt} for examples. To increase the diversity and frequency of transparent and reflective materials, we modify the scenes by randomly modifying scene lighting and materials, and also by randomly placing additional non-Lambertian objects like glass bottles and windows. Per-pixel ground truth is generated via ray tracing.
Experiments show that fine-tuning on our synthetic dataset indeed helps existing optical flow methods achieve better single-layer results on the benchmark, especially for the non-Lambertian surfaces. We further offer a baseline method for multi-layer optical flow based on RAFT \cite{raft}, providing a starting point for multi-layer non-Lambertian object perception.

In summary, our contributions are as follows:
\begin{itemize}
    \item We provide a diverse real world multi-layer benchmark for non-Lambertian objects with 150k high quality optical flow and stereo pairs in 185 scenes, and evaluate state-of-the-art optical flow methods on it.
    \item We provide a large-scale synthetic dataset for non-Lambertian objects with 60k images in 30 scenes, enriched with random placement of objects and random alternation of materials. Fine-tuning on the dataset helps existing optical flow methods achieve better single-layer results on the benchmark.
    \item We propose a novel task, multi-layer optical flow estimation and offer a RAFT-based baseline method.
\end{itemize}

\section{Related Work}
\label{sec:related}
\subsection{Non-Lambertian Benchmarks and Datasets}
\paragraph{Synthetic Non-Lambertian Datasets}have been introduced in previous works \cite{cleargrasp, lit, zhu2021rgb} to serve as training data for downstream vision tasks. While it has been shown that models trained on synthetic data can generalize to real world scenarios, synthetic data cannot serve as a satisfactory benchmark due to the visual gap between synthetic and real world images. In contrast, we collect 2000 real world images of 185 indoor and outdoor scenes and 360 different objects to form a comprehensive non-Lambertian benchmark.

\paragraph{Real-World Non-Lambertian Benchmarks}exist for 2D tasks such as segmentation \cite{transcut, trans10k, donthit, richcontext} and image matting \cite{tom-net}. While these are useful tasks, our benchmark is geared towards 3D tasks, as we include multi-layer optical flow annotations and camera poses. Other real world benchmarks have been developed for 3D tasks such as pose estimation \cite{stereobj1m, keypose, seeingglassware} and depth estimation \cite{phocal, clearpose, transcg, seeingglass}. However, the techniques these works use to acquire accurate 3D information for non-Lambertian objects greatly limits the scene and object diversity of their benchmarks. Some works align 3D models of pre-scanned non-Lambertian objects with the images they appear in. But this limits the objects used to those that can be 3D scanned, and scenes are often constrained to small objects placed on a desk. Booster \cite{booster} paints non-Lambertian objects with Lambertian paint and projects random patterns onto the paint to aid ground truth stereo computation. However, this procedure is hard to scale because it requires intensive manual labor, and is also limited to indoor scenes due to the use of structured lighting. Liang \etal \cite{glasswall} sparsely paste opaque covers on glass walls and interpolate the measurements to derive ground truth 3D, but this limits the approach to planar surfaces. Our benchmark does not face these limitations of scene and object diversity, as the AprilTag system can be used in most scenes and can be applied to non-Lambertian objects of most scales. See Tab.~\ref{tab:related_work} for a detailed comparison of our benchmark to these works.

\paragraph{See-Through Methods and Benchmarks} such as \cite{qiu2023looking, seeing_through_the_glass} focus on the task of predicting 3D geometry of objects placed behind glass or other highly specular and transparent surfaces. However, these works focus solely on evaluating geometry of diffuse objects behind the initial transparent surface, and also only use one layer of transparent material for occlusion. In contrast, our benchmark provides ground truth measurements on layers including the initial transparent layer , and we also consider cases where multiple layers of transparent surfaces are present.

\begin{table*}[t]
  \centering
  \resizebox{\linewidth}{!}{
  \begin{tabular}{l|c|c|c|c|c|c|c|c}
    \toprule
        \multirow{2}{*}{Benchmark} & \multirow{2}{*}{Domain} & Multiple & \multirow{2}{*}{Depth} & \multirow{2}{*}{Disparity} & \multirow{2}{*}{Flow} & \# Non-Lambertian & \# Scenes & \# Real Frames  \\
        & & Layer & & & & Objects in Total & in Total & in Total \\
    \midrule 
        ClearGrasp \cite{cleargrasp} & Indoor & & \text{\ding{51}} & & & 10 & 25 & 286 \\
        ClearPose \cite{clearpose} & Indoor & & \text{\ding{51}}& & &  63 & 51 & 350K  \\
        TransCG \cite{transcg} & Indoor & & \text{\ding{51}} & & &  51 & 130 & 57715  \\
        TODD \cite{seeingglass} & Indoor & & \text{\ding{51}} & & &  6 & 5 & 14659 \\
        PhoCal \cite{phocal} & Indoor & & \text{\ding{51}} & & &  25 & 24 & 7118  \\
        Booster \cite{booster} & Indoor & & \text{\ding{51}} & \text{\ding{51}} & & 108 & 64 & 419 \\
        Liang \etal \cite{glasswall} & Indoor, Outdoor & & \text{\ding{51}} & & & 66 & 66 & 1200 \\
        Ours & Indoor, Outdoor & \text{\ding{51}} & \text{\ding{51}} & \text{\ding{51}} & \text{\ding{51}} & 360 & 185 & 2000 \\
    \bottomrule
  \end{tabular}}
  \caption{Comparison of real world non-Lambertian objects benchmarks. Our benchmark exhibits greater scene and object diversity, and is the first to provide multi-layer optical flow annotation.}
  \label{tab:related_work}
\end{table*}

\subsection{Optical Flow Datasets}
\paragraph{Synthetic Optical Flow Datasets.} A large number of synthetic optical flow datasets have been proposed \cite{sintel, spring, flownet, mayer2016large, virtual-kitti, human-optical-flow, richter2017playing, autoflow}, but these are not sufficient to serve as benchmarks due to the sim-to-real gap. Our benchmark consists of real world ground truth optical flow and stereo pair annotations.

\paragraph{Real-World Optical Flow Benchmarks.} are much rarer compared to their synthetic dataset counterparts. Notable examples include \cite{kitti1, kitti2, hd1k}, but they only capture data from automotive scenes. Our benchmark covers a greater variety of scenes and objects. In addition, the techniques employed by these existing benchmarks such as structured lighting \cite{middleburry} and time-of-flight \cite{kitti1, eth3d, hd1k} struggle to accurately capture characteristics of non-Lambertian objects, and are unable to capture ground truth data behind transparent objects. In contrast, our data acquisition pipeline is able to effectively deal with these challenging cases.

\vspace{-1mm}
\paragraph{Optical Flow Methods.} Optical flow has been a long-standing fundamental task in the field of computer vision. Strategies for this problem include optimizing for visual similarity \cite{black1993framework, bruhn2005lucas, farneback2003two, horn1981determining}, using deep learning and CNNs \cite{flownet}, performing iterative refinement \cite{raft, gma, jiang2021learning, skflow}, using transformer-based architectures for feature matching \cite{transformer, gmflow, gmflow+, croco_v2, flowformer, flowformer++}, and using coarse-to-fine stretgies for handling large displacements \cite{spynet, liteflownet, pwc, yang2019volumetric, hofinger2020improving}. However, none of these works solve the problem setting of predicting optical flow on multiple layers. On the other hand, our proposed baseline, which is trained on our comprehensive multi-layer optical flow synthetic dataset, is able to produce multi-layer predictions.

\section{Collecting \projectname{}'s Real World Benchmark}
\label{sec:benchmark}

Through our real world dataset benchmark, we aim to achieve three main goals. First, we wish to capture highly accurate optical flow ground truth for non-Lambertian objects. Second, in the presence of transparent objects, we wish to record multi-layer information. Finally, we wish to have high diversity of objects and scenes within the dataset.
Historically, these goals have been hard to achieve through frequently employed 3D ground truth acquisition methods. In this section, we describe how we utilize AprilTag \cite{apriltags}, a visual fiducial system, within a stereo camera system to overcome these challenges.

\subsection{Introduction to AprilTag}

AprilTag \cite{apriltags} is a visual fiducial marker system used to collect precise 3D positions and orientations of real world objects. The system consists of two components: AprilTag markers, which are easily-identifiable 2D bar code-style tags, and AprilTag detection software which conducts the desired measurements. 

In our approach, AprilTag markers are printed in various sizes on matte vinyl stickers, which can be easily peeled off surfaces without leaving any residue. By capturing images both with and without the AprilTag markers in place, we are able to generate stereo and motion image pairs along with corresponding multi-layer ground truth measurements.

\subsection{Data Acquisition and Annotation}

\paragraph{Camera Setup and Calibration.}
Our real world images are captured by two 4k webcams fixed to a dual camera mount on top of a tripod. Each image is captured at a resolution of $3840 \times 2160$. Our benchmark contains a wide range of object sizes, from small items like cups and tubes to large items like cars or glass walls. As a result, it is necessary to adjust the focal length of our cameras to guarantee that the subjects are in focus. Before each image capture, we manually select appropriate focal lengths to keep the object of interest in focus, and we perform both single-camera and stereo calibrations using OpenCV \cite{opencv}.

\paragraph{Image Capturing Pipeline.}

\begin{figure}[t]
    \centering
    \includegraphics[width=0.8\linewidth]{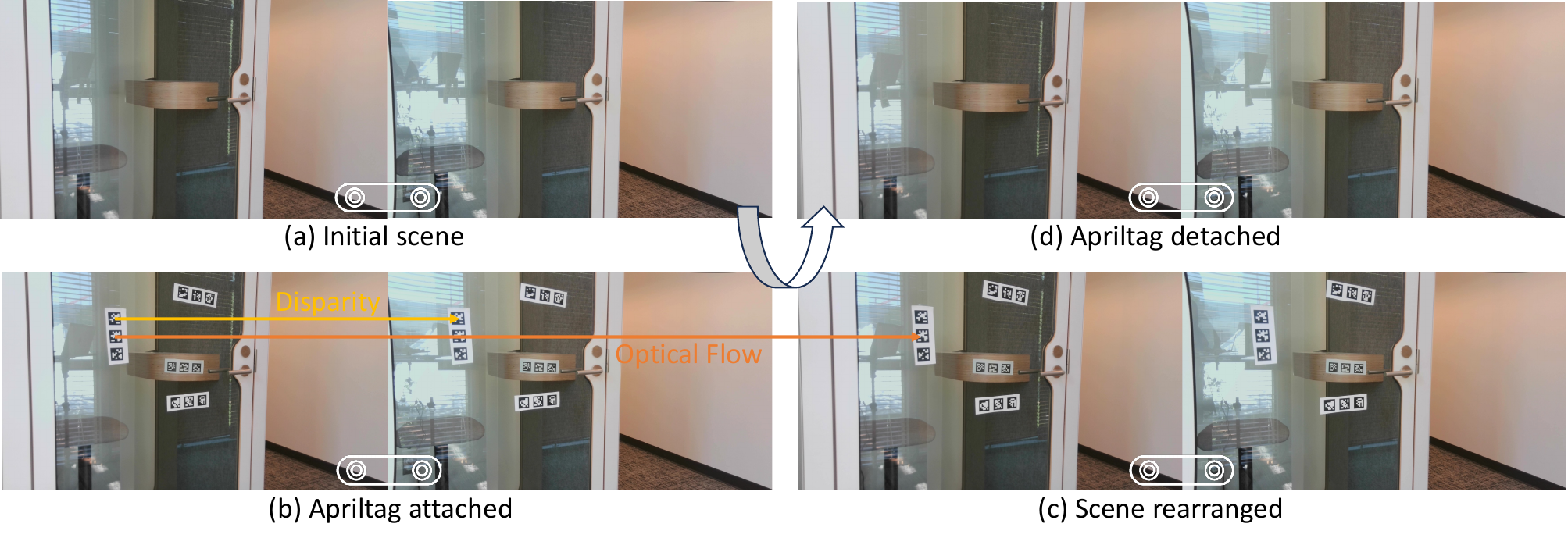}
    \caption{Image Capturing Pipeline. (a) The original stereo image pair. (b) AprilTags are carefully attached to the scene, allowing disparity measurements (yellow arrow). (c) The scene is altered by changing the positions and orientations of scene objects and the stereo camera system. Optical flow is measured via AprilTags (orange arrow). (d) AprilTags are detached, yielding the final tagless stereo image pair.}
    \label{fig:pipeline}
\end{figure}

For each object in the benchmark, our image capturing pipeline consists of four steps which correspond to \cref{fig:pipeline}(a)-(d). Each object's image set consists of eight images: two stereo image pairs of the scene without AprilTag markers, and two stereo image pairs with AprilTags that help provide ground truth measurements.

First, we deploy our calibrated stereo camera system to capture the first pair of images (a) without AprilTag markers. Next, we carefully attach AprilTag markers to scene objects to avoid altering object placements and capture our next stereo pair (b). Each tag, associated with a unique identifier, provides stereo correspondence for the initial image pair. We then re-arrange the scene by changing the location and orientation of movable objects and the cameras; this allows us to take the next stereo pair (c) with AprilTags still attached. Finally, we carefully remove AprilTag markers and take the last unmarked image pair (d). By using the AprilTag measurements from (b) and (c), we can compute ground truth for optical flow for the unmarked image pairs (a) and (d).

\paragraph{Multi-Layer Ground Truth.}
\label{para:multilayer_gt}

Traditionally, each pixel in an image corresponds to the first object or surface its corresponding camera ray intersects in a scene. However, when transparent objects are present in images, camera ray intersections with objects behind the first object may be visible as well. Specifically, pixels may contain information about multiple layers, where each layer represents a point at which the pixel's camera ray makes contact with the surface of a visible object. Accurately perceiving information of layers is challenging due to the complexities of light reflection and refraction at the interface between air and transparent surfaces.

Our data collection method's flexibility allows the placement of AprilTag markers behind transparent surfaces, effectively capturing ground truth while preserving inherent refraction and distortion effects. As shown in \cref{fig:pipeline}, markers placed on the desk behind the glass door remain visible and easy to detect. Remarkably, even under extreme refraction conditions, many markers are still identifiable and provide reliable ground truth. This enables the acquisition of ground truth data from various depth layers, thus preserving the effects of light refraction and producing annotations aligned with human perception.

\paragraph{Postprocessing and Annotation.}

Using our stereo camera calibration, we rectify the stereo images and conduct point triangulation to obtain 3D position of each marker. While the primary focus of this paper is 2D correspondences, our benchmark is also capable of serving as a benchmark for depth and scene flow. We also manually annotate each marker with its material property and layer index.

\subsection{Benchmark Statistics}

The flexibility of our data collection procedure allows us to collect data from a diverse set of scenes. We captured 1000 stereo image pairs with optical flow, including 400 validation and 600 test scenes. The dataset contains 360 distinct objects placed in 155 indoor and 30 outdoor scenes under different lighting conditions. Objects include common indoor elements such as glass walls, doors, and staircases; household items like knives, sinks, pots, and washing machines; laboratory equipment like robots, printers, beakers, and tubes; and outdoor items like cars, fire hydrants, and bus stop shelters.

The number of AprilTag markers placed in each scene determines how many correspondence ground truth provided by each image pair, varying from 20 to 500 correspondences. Overall, we provide approximately 150k correspondences for both stereo and optical flow. Detailed statistics can be found in supplementary.

\section{Synthetic Dataset}
\label{sec:dataset}

\begin{figure*}[tb]
    \centering
    \includegraphics[width=0.8\linewidth]{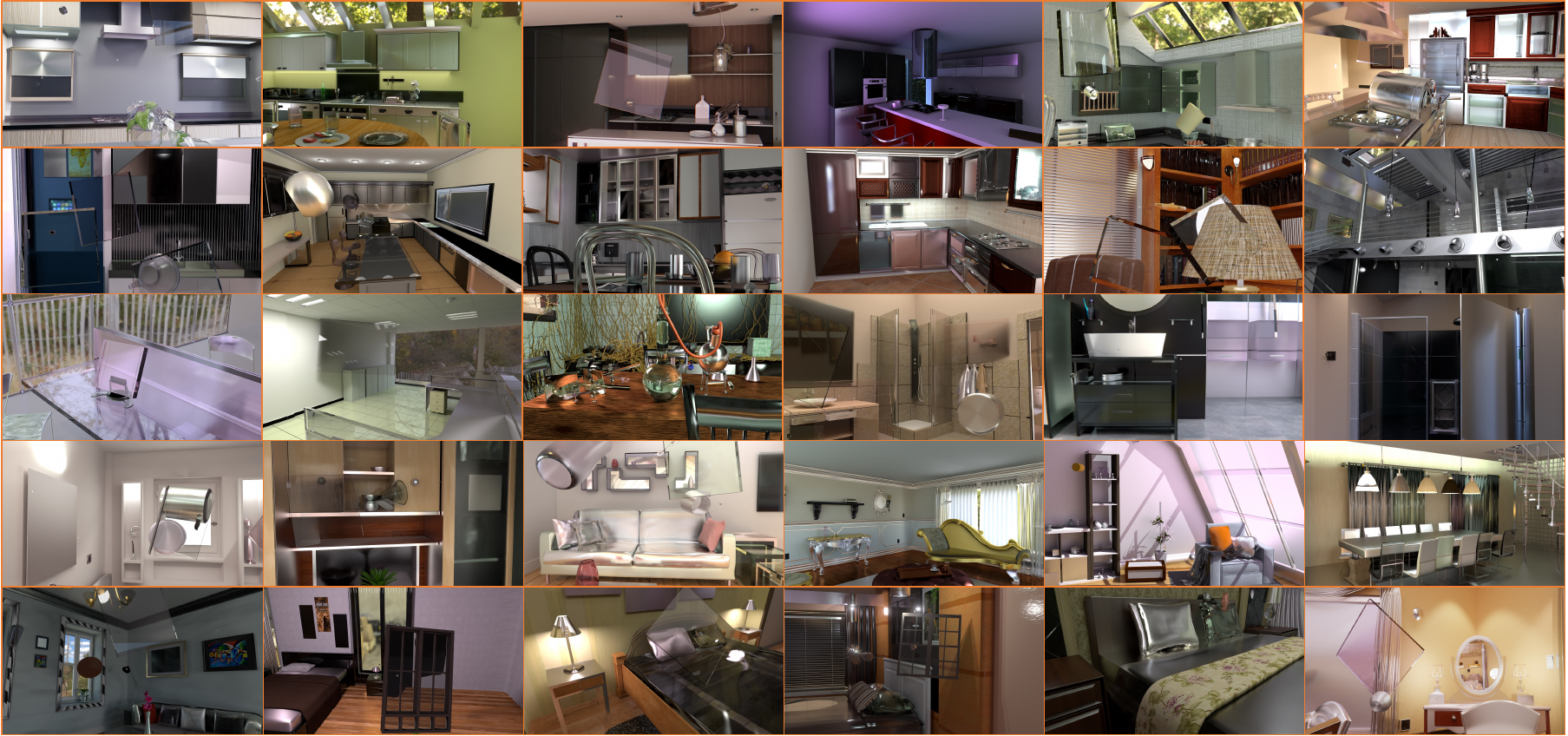}
    \caption{Gallery of our synthetic dataset. Our synthetic dataset is generated from modified versions of 30 high-quality indoor scenes designed by artists. To increase the frequency of non-Lambertian objects and diversity of images, we randomly modify material properties, change scene lighting, and insert additional objects.}
    \label{fig:synthetic_gallery}
\end{figure*}

Due to the inherent challenges in annotating non-Lambertian objects, acquiring a real world dataset with multi-layer ground truth annotations that are sufficiently dense for training purposes is improbable. To address this limitation, we created a comprehensive synthetic dataset of 60k images of non-Lambertian objects via Blender \cite{blender}. These images are rendered from modified versions of 30 high-quality indoor BlendSwap\cite{BlendSwap} scenes designed by artists, including 10 kitchens, 5 bathrooms, 5 offices, 5 living rooms and 5 bedrooms, shown in \cref{fig:synthetic_gallery}.

\begin{figure*}[t]
    \centering
    \includegraphics[width=0.8\linewidth]{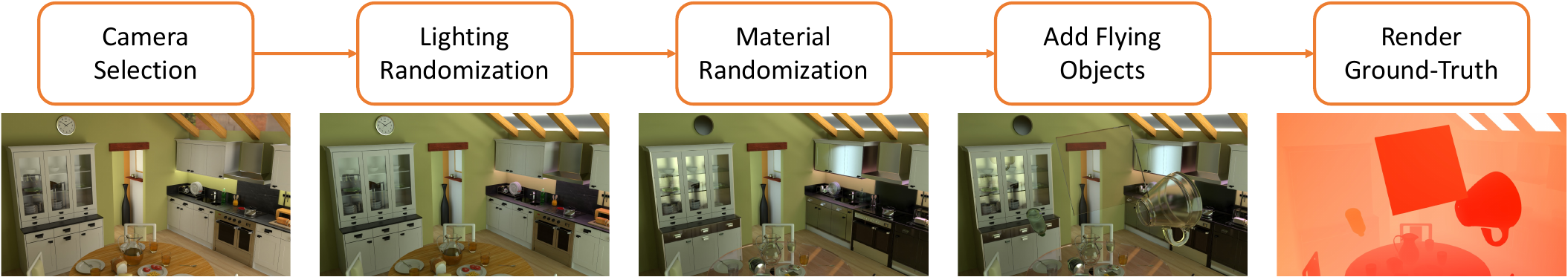}
    \caption{Synthetic dataset generation pipeline. We perform these steps to boost the frequency of non-Lambertian object appearances and the diversity of rendered images.}
    \label{fig:synthetic_pipeline}
\end{figure*}

Although the original BlendSwap scenes contain some non-Lambertian objects, these objects may not be in view at all viewpoints. To boost the frequency of non-Lambertian objects and the diversity of collected images, we utilize the following image generation process which is done via Blender Python API. See \cref{fig:synthetic_pipeline} for an illustration of this process.

\begin{itemize}
    \item \emph{Camera Selection.} A camera can be defined by its position, orientation, and focal length. For each image, we randomly select camera settings from a manually specified subset of parameter combinations for each scene. By carefully pruning the parameter space, we preserve viewpoint diversity while avoiding trivial images, such as views of a blank wall.
    \item \emph{Lighting Randomization.} We randomly assign colors and intensities to all light sources in the scene. We further modify the environment textures, randomly selecting from 50 different HDR images sourced from HDRi Haven\cite{HDRI-haven}.
    \item \emph{Material Randomization.} We randomly select some objects and alter their material properties to be glass or metal with varying color and roughness.
    \item \emph{Add Flying Objects.} We add several random objects to scene, sampled from 100 distinct BlendSwap object categories including bottles, pots, sculptures, windows, etc. These objects also are randomly assigned material properties and are randomly placed to be in the field of view of the camera.
\end{itemize}

We also provide multi-layer ground truth 3D annotations, as shown in \cref{fig:synthetic_gt}. These annotations are visually aligned with the camera's view of the scene, as we take into account visual distortions due to refraction as opposed to directly projecting objects' ground truth positions onto the imaging plane. We modified Blender's ray tracing source code to embed our ground truth collection in the rendering process. During ground truth rendering, we only consider materials with sufficiently low roughness value to be transparent---this threshold is set to make sure we provide multi-layer data only for surfaces that appear transparent to human observers. To ensure we only track rays emanating from real objects instead of their reflections, we disable reflective rays in the scene. We determine the layer corresponding to each ray by tracking the number of transparent surfaces it hits. By aggregating all this information, we generate ground truth data that is consistent with human perception of multi-layer scenes.

\section{Multi-layer Optical Flow Task and Baseline}

\subsection{Problem Formulation}
\label{subsec:multilayer_problem}

Given two images $\mathcal{I}_1$ and $\mathcal{I}_2$ with dimensions $H \times W \times 3$ and a query pixel $p = (x, y)$ as input, the goal of multi-layer optical flow is to produce a sequence of ordered per-layer optical flow predictions $\hat{\mathcal{F}} = \{\hat{\mathbf{f}}_1, ..., \hat{\mathbf{f}}_n\}$, where $n$ can be chosen by the model to vary with respect to the query pixel. Each layer's optical flow  prediction $\hat{\mathbf{f}}_i$ is a 2D vector which represents the displacement of the $i$th layer of surface from pixel $(x, y)$ of $\mathcal{I}_1$ to its corresponding pixel in $\mathcal{I}_2$.

\subsection{Evaluation Metrics}
\label{sec:multilayer_metric}

Consider a pixel $p = (x, y)$ with ground truth optical flow annotations on $k$ arbitrary layers, with layer indices $m_1 \leq m_2 \leq ... \leq m_k$. Let $\mathcal{F}=\{\mathbf{f}_{m_1}, ...,\mathbf{f}_{m_k}\}$ denote the ground truth optical flow vectors, and let $\{t_{m_1}, ..., t_{m_k} \}$ be ground truth transparent material indicators where $t_{m_i} = 1$ if the material of layer $m_i$ is transparent and $t_{m_i} = 0$ otherwise. Then, given $n$ flow predictions $\hat{\mathcal{F}} = \{\hat{\mathbf{f}}_1, ..., \hat{\mathbf{f}}_n\}$ for pixel $p$, we consider the following per-pixel criteria:

\textbf{Layer Count Correctness} denotes whether the predicted number of layers $|\hat{\mathcal{F}}|$ for pixel $p$ is plausible with respect to $p$'s ground truth annotations.
If the last ground truth layer $m_k$ is annotated as transparent, then there should be at least $m_k$ predicted layers. Conversely, if layer $m_k$ contains non-transparent material, no layer can exist behind it and so the number of predicted layers should equal $m_k$. Thus the predicted layer count of a pixel is correct only if
\begin{align*}
\begin{cases}
 |\hat{\mathcal{F}}| \geq m_k, & \text{when } t_{m_k} = 1\\
|\hat{\mathcal{F}}| = m_k & \text{when } t_{m_k} = 0
\end{cases}
\end{align*}

\textbf{Flow Prediction $\tau$-Accuracy} denotes whether the predicted optical flows for pixel $p$ are all within an L2 radius of $\tau$ of their corresponding ground truth optical flows, if known. Specifically, when $n \ge m_k$, pixel $p$'s multi-layer flow prediction is considered $\tau$-accurate if
\begin{align*}
\max_i || \hat{\mathbf{f}}_{m_i} - \mathbf{f}_{m_i} ||_2 \leq \tau
\end{align*}

In our benchmark, each annotated pixel is annotated with one layer $l$ of optical flow ground truth along with the material. So the flow prediction $\tau$-accuracy reduces to whether the following inequality holds:
\begin{align*}
    || \hat{\mathbf{f}}_l - \mathbf{f}_l ||_2 \leq \tau
\end{align*}

Using these two criteria, we define the following evaluation metrics for predicted multi-layer optical flow over all pixels in $\mathcal{I}_1$: 1) \textbf{Multi-Layer bad-$\tau$} denotes the percentage of pixels with flow prediction that is not $\tau$-accurate; and 2) \textbf{Multi-Layer Count-Aware bad-$\tau$} denotes the percentage of pixels with either incorrect layer count or flow prediction that is not $\tau$-accurate. 

\subsection{Baseline Method Design}

Taking inspiration from RAFT\cite{raft}, our baseline method contains three main parts: a feature encoder that extracts per-pixel features from both input images to construct a 4D correlation volume, context encoders that extract features from only the first image, and an update operator which recurrently updates optical flow. Unlike RAFT, we utilize $n$ context encoders instead of 1 to separately extract context features for each layer. Each context encoder shares the same architecture but has independent weights. The $n$ outputs are then separately fed into a ConvGRU-based update block to produce $n$ optical flow predictions $\hat{\mathcal{F}}$.

During training, for a training sample provides $k$ layers of optical flow ground truth, we duplicate the last layer $n-k$ times to meet the $n$ predictions and utilize the common optical flow training loss \cite{raft} for each layer. This approach can apply to both our multi-layer synthetic dataset and existing single-layer datasets.
During inference, after obtaining $n$ raw optical flow predictions, we perform pruning heuristics to avoid repetitive predictions. For each layer, we discard flow predictions that are within a radius of $\delta$ pixels of the previous layer's corresponding flow prediction. Performing this pruning simultaneously for each layer produces final prediction $\hat{\mathcal{F}}$. We set $n=4, \delta = 0.5$ for our baseline.

\section{Experiments}
\label{sec:experiments}

\subsection{Single-Layer Experiments}

To evaluate effectiveness of existing methods and to show the usefulness of our synthetic dataset for non-Lambertian optical flow, we compare the performance of RAFT finetuned on our dataset against representative optical flow models on the benchmark. 
Our multi-layer benchmark serves as a single-layer optical flow benchmark here by limiting evaluation to points on a \emph{single layer} subset.

\paragraph{Metrics and Implementation Details.} We adopt the commonly-used average end-point-error (EPE) and single-layer bad-$\tau$ metrics. EPE measures the average L2 distance between predicted and ground truth optical flow. Bad-$\tau$ represents the percentage of pixels having L2 error larger than a threshold of $\tau$. Evaluation is done on \projectname{} with images downsampled to a resolution of $540 \times 960$ due to memory constraints.
For existing optical flow models, we directly use their publicly available implementations \cite{ptlflow}. We use pre-trained weights fine-tuned for Sintel \cite{sintel} if accessible; otherwise, we use FlyingThings3D \cite{mayer2016large} weights. Test-time optimizations such as tiling technique are disabled for a fair comparison. For fine-tuning RAFT, we start with its pre-trained weights for Sintel. From here, we employ one of three fine-tuning approaches: 1) L: directly fine-tuning on our synthetic dataset; 2) S: directly fine-tuning on Sintel; and 3) S+L: jointly fine-tuning on both our synthetic dataset and Sintel. We perform fine-tuning and evaluation on a single NVIDIA RTX 3090 GPU.

\paragraph{First Layer Optical Flow Experiments.} 
Under this setting, we only evaluate methods' optical flow predictions for \projectname{}'s first layer points. 
For transparent objects, methods must predict the optical flow of the transparent occluder rather than the background, similar to setting of other non-Lambertian benchmarks \cite{booster, phocal}. We fine-tune RAFT on first layer points in our synthetic dataset.

We report evaluation results for all first layer points and material-specific subcategories, as shown in \cref{tab:optical_flow_first_layer}. 
All methods incur significantly greater error on our benchmark compared to performances on other optical flow benchmarks \cite{sintel, kitti1, kitti2}, proving the challenging nature of our benchmark.
Overall, fine-tuning RAFT with our synthetic dataset boosts the model's performance on non-Lambertian surfaces. Jointly fine-tuning with Sintel (S+L) makes the performance more stable and better maintains the performance on diffuse points, thereby delivering the best results. The S+L fine-tuned RAFT reduces the overall EPE and bad-$\tau$ for all non-Lambertian surfaces, while fine-tuning solely on Sintel does not yield any improvements, proving the usefulness of our synthetic dataset.

\begin{table*}[h]
  \centering
  \resizebox{\linewidth}{!}{
  \begin{tabular}{l c cccc c cccc c cccc c cccc}
    \toprule
    \multirow{2}{*}{Method} && \multicolumn{4}{c}{All} && \multicolumn{4}{c}{Transparent} && \multicolumn{4}{c}{Reflective} && \multicolumn{4}{c}{Diffuse} \\
    \cmidrule{2-21}
    && EPE$\downarrow$ & 1px$\downarrow$ & 3px$\downarrow$ & 5px$\downarrow$ 
    && EPE$\downarrow$ & 1px$\downarrow$ & 3px$\downarrow$ & 5px$\downarrow$ 
    && EPE$\downarrow$ & 1px$\downarrow$ & 3px$\downarrow$ & 5px$\downarrow$
    && EPE$\downarrow$ & 1px$\downarrow$ & 3px$\downarrow$ & 5px$\downarrow$\\
    \midrule
    FlowNet-C \cite{flownet} && 21.14 & 94.88 & 77.86 & 65.20 && 24.01 & 94.84 & 77.90 & 65.15 && 13.85 & 94.82 & 79.18 & 67.35 && 17.04 & 96.18 & 70.95 & 56.46 \\
    FlowNet2 \cite{flownet2} && 20.67 & 86.42 & 66.54 & 56.66 && 23.54 & 87.19 & 67.61 & 57.55 && 13.52 & 84.57 & 63.45 & 53.97 && 15.42 & 76.30 & 54.82 & 47.54 \\
    PWC-Net \cite{pwc} && 28.39 & 83.93 & 63.66 & 54.33 && 31.75 & 86.34 & 66.69 & 57.10 && 15.45 & 74.12 & 51.02 & 43.56 && 20.48 & 70.80 & 48.74 & 37.06 \\
    GMA \cite{gma} && 16.58 & 79.26 & 57.04 & 46.60 && 20.35 & 82.93 & 61.16 & 49.91 && \textbf{8.18} & 65.34 & 41.04 & 33.83 && 12.00 & 55.04 & 31.45 & 25.48 \\
    SKFlow \cite{skflow} && 18.14 & 79.12 & 57.47 & 48.33 && 22.17 & 83.31 & 62.01 & 52.12 && 9.41 & 62.38 & \textbf{38.95} & 32.89 && 8.17 & 55.15 & 33.09 & 28.21 \\
    CRAFT \cite{craft} && 17.82 & 80.31 & 57.60 & 47.90 && 21.57 & 84.07 & 61.86 & 51.34 && 10.11 & 64.79 & 40.48 & 33.91 && 8.73 & 60.94 & 33.78 & 29.49 \\
    GMFlow \cite{gmflow} && 16.92 & 88.45 & 64.00 & 51.71 && 20.72 & 89.51 & 65.90 & 54.01 && 8.74 & 85.86 & 58.01 & 43.18 && 8.29 & 74.63 & 45.58 & 35.64  \\
    GMFlow+ \cite{gmflow+} && 17.62 & 89.83 & 67.21 & 54.29 && 21.36 & 90.36 & 68.83 & 56.45  && 9.68 & 88.65 & 61.91 & 45.80 && 10.06 & 82.53 & 52.35 & 41.21 \\
    FlowFormer \cite{flowformer} && 18.49 & 78.83 & 58.61 & 49.24 && 22.56 & 83.02 & 63.42 & 53.64 && 9.54 & \textbf{61.73} & 39.63 & \textbf{32.24} && \textbf{5.01} & 56.57 & 30.21 & \textbf{21.33} \\
    RAFT \cite{raft} && 16.49 & 78.45 & 55.64 & 45.78 && 20.11 & 82.72 & 59.69 & 49.06 && 8.51 & 62.05 & 40.49 & 33.21 && 10.76 & \textbf{50.78} & \textbf{27.56} & 24.39 \\
    \midrule
    RAFT-ft. (S) && 17.94 & 79.53 & 59.47 & 49.69 && 21.96 & 82.94 & 63.15 & 52.85 && 8.89 & 66.41 & 45.21 & 37.11 && 9.07 & 57.70 & 36.44 & 31.34 \\
    RAFT-ft. (L) && 17.46 & \underline{78.13} & \underline{53.12} & \underline{43.33} && \underline{18.54} & \underline{82.15} & \underline{\textbf{56.06}} & \underline{45.73} && 
    17.30 & 62.60 & 41.75 & 33.89 && 14.69 & 52.60 & 34.73 & 28.87 \\
    RAFT-ft. (S+L) && \underline{\textbf{15.63}} & \underline{\textbf{77.81}} & \underline{\textbf{52.75}} & \underline{\textbf{42.76}}
                   && \underline{\textbf{18.39}} & \underline{\textbf{81.88}} & \underline{56.17} & \underline{\textbf{45.40}}
                   && 11.73 & \underline{61.93} & \underline{39.48} & \underline{32.97}
                   && \underline{6.95} & 52.75 & 31.23 & \underline{24.24} \\
    \bottomrule
  \end{tabular}
  }
  \caption{Representative optical flow methods evaluated on first layer subset of our benchmark using EPE and bad-$\tau$ metrics. Best scores are in \textbf{bold}. \underline{Underlined numbers} denote RAFT fine-tuned on our synthetic data outperforming the original version.}
  \label{tab:optical_flow_first_layer}
\end{table*}

\begin{table*}[t]
  \centering
  \resizebox{\linewidth}{!}{
  \begin{tabular}{l c cccc c cccc c cccc c cccc}
    \toprule
    \multirow{2}{*}{Method} && \multicolumn{4}{c}{All} && \multicolumn{4}{c}{Reflective} && \multicolumn{4}{c}{Diffuse} && \multicolumn{4}{c}{Behind Transparent} \\
    \cmidrule{2-21}
    && EPE$\downarrow$ & 1px$\downarrow$ & 3px$\downarrow$ & 5px$\downarrow$ 
    && EPE$\downarrow$ & 1px$\downarrow$ & 3px$\downarrow$ & 5px$\downarrow$ 
    && EPE$\downarrow$ & 1px$\downarrow$ & 3px$\downarrow$ & 5px$\downarrow$
    && EPE$\downarrow$ & 1px$\downarrow$ & 3px$\downarrow$ & 5px$\downarrow$ \\
    \midrule
    FlowNet-C \cite{flownet} && 14.94 & 94.87 & 75.44 & 61.66 && 13.54 & 94.89 & 78.26 & 65.62 && 17.13 & 94.82 & 68.83 & 52.36 && 16.37 & 93.95 & 64.60 & 43.73 \\
    FlowNet2 \cite{flownet2} && 14.24 & 80.88 & 58.87 & 48.59 && 13.21 & 84.79 & 63.42 & 52.45 && 14.98 & 71.70 & 48.18 & 39.54 && 13.67 & 70.28 & 44.35 & 28.40 \\
    PWC-Net \cite{pwc} && 19.39 & 72.59 & 48.44 & 40.10 && 14.98 & 74.06 & 50.42 & 42.20 && 23.78 & 69.15 & 43.79 & 35.11 && 24.87 & 68.12 & 38.06 & 29.10 \\
    GMA \cite{gma} && \textbf{7.55} & 62.00 & 35.99 & 29.13 && \textbf{7.95} & 65.12 & 40.63 & 32.77 && 9.51 & 54.67 & 25.11 & 20.56 && 6.81 & 54.81 & 20.09 & 13.80 \\
    SKFlow \cite{skflow} && 8.04 & 59.83 & 34.78 & 29.29 && 9.14 & 62.37 & 38.13 & 31.93 && 7.20 & 53.86 & 26.92 & 23.10 && 7.38 & 53.80 & 19.88 & 16.11 \\
    CRAFT \cite{craft} && 8.25 & 62.43 & 36.66 & 30.27 && 9.82 & 64.59 & 40.21 & 33.22 && 7.66 & 57.36 & 28.32 & 23.36 && 6.87 & 54.48 & 24.14 & 16.70 \\
    GMFlow \cite{gmflow} && 7.69 & 82.16 & 52.94 & 39.09 && 8.52 & 85.53 & 57.77 & 42.16 && 7.11 & 74.24 & 41.60 & 31.89 && 6.15 & 74.07 & 39.32 & 26.02 \\
    GMFlow+ \cite{gmflow+} && 8.52 & 86.11 & 57.60 & 42.52 && 9.46 & 88.76 & 62.08 & 45.13 && 7.64 & 79.89 & 47.09 & 36.40 && 5.81 & 79.23 & 45.19 & 30.77 \\
    FlowFormer \cite{flowformer} && 7.96 & 59.65 & 35.40 & 27.64 && 9.28 & 62.17 & 39.66 & 31.50 && \textbf{6.26} & 53.74 & 25.40 & \textbf{18.57} && 8.12 & 54.14 & 23.24 & 14.94 \\
    RAFT \cite{raft} && 8.28 & 58.68 & 34.96 & 28.86 && 8.25 & 61.68 & 39.38 & 32.18 && 9.58 & \textbf{51.66} & \textbf{24.57} & 21.06 && 8.48 & 52.19 & 19.54 & 15.59 \\
    \midrule
    RAFT-ft. (L)  && \underline{7.72} & \underline{\textbf{57.67}} & \underline{\textbf{34.32}} & \underline{\textbf{27.15}} && 9.43 & \underline{\textbf{59.71}} & \underline{\textbf{37.09}} & \underline{\textbf{29.20}} && \underline{7.79} & 52.88 & 27.80 & 22.33 && \underline{\textbf{4.34}} & \underline{\textbf{52.07}} & \underline{\textbf{19.01}} & \underline{\textbf{12.90}} \\
    \bottomrule
  \end{tabular}}
  \caption{Representative optical flow methods evaluated on last layer subset of our benchmark using EPE and bad-$\tau$ metrics. Best scores are in \textbf{bold}. \underline{Underlined numbers} denote RAFT fine-tuned on our synthetic data outperforming the original version.}
  \label{tab:optical_flow_last_layer}
\end{table*}

\paragraph{Last Layer Optical Flow Experiments.} 

For this set of experiments, we only evaluate methods' optical flow predictions for \projectname{}'s last layer points. To guarantee the points are on the last layer, we only consider points that are associated with non-transparent materials. 
In addition to reporting results for all last layer points and material-specific subcategories, we also consider a new category called \textit{Behind Transparent}. This category contains last layer points that are behind at least one transparent layer, testing the capacity of the methods to see through transparent surfaces, bearing resemblance to the ``see-through'' problem \cite{qiu2023looking, seeing_through_the_glass}. 
We fine-tune RAFT on last layer points in our synthetic dataset.

Results are shown in \cref{tab:optical_flow_last_layer}. 
Compared to first layer experiment errors, the errors for last layer are significantly smaller, demonstrating that first layer optical flow estimation is a challenging problem due to the misalignment between visual appearance and 3D geometry. When transparent objects are present, existing methods often see through and fail to consider their structure.
Overall, fine-tuned RAFT outperforms the original version, reducing the overall EPE from 8.28 to 7.72 and all bad-$\tau$ for all non-Lambertian surfaces. The improvement on the Behind Transparent category is particularly notable---fine-tuning reduces EPE from 8.48 to 4.34.
These results highlight the effectiveness of our synthetic dataset, especially in its ability to help enhance models' ability to see through specular and reflective effects.
In conclusion, our synthetic dataset greatly aids data-driven optical flow techniques in handling non-Lambertian objects. See our supplementary material for additional single-layer experiments.

\subsection{Multi-Layer Baseline Evaluation on \projectname{}}
\label{sec:multi_layer}

In this section, we evaluate our baseline method, which essentially is a multi-headed RAFT-base architecture (Multi-RAFT). Similar to how we fine-tune RAFT on our synthetic dataset, we have several strategies to train our baseline method. First, each head's weights are initialized to pre-trained Sintel weights, and we experiment with fine-tuning all heads with either the L or S+L policy.

We compare our baseline model Multi-RAFT to the original RAFT \cite{raft}, and we evaluate using multi-layer count-aware bad-$\tau$ with $\tau = 1, 3, 5, \infty$. Note that $\tau = \infty$ corresponds to the degenerate case of only caring about pixel layer prediction accuracy. As RAFT only provides single optical flow prediction, the number of layers predicted $|\hat{\mathcal{F}}|$ is always 1, which means it will automatically get 0\% error rate on $\tau=\infty$ setting of Layer 1. However, Multi-RAFT beats RAFT on all other metrics on first layer, proving the strength of our synthetic dataset and our baseline model design. We categorize ground truth points in our benchmark by the layer they are on and show evaluation results in \cref{tab:evaluation_multilayer}.

\begin{table*}[t]
  \centering
  \vspace{-3mm}
  \resizebox{\linewidth}{!}{
  \begin{tabular}{l c cccc c cccc c cccc}
    \toprule
    \multirow{2}{*}{Method} && \multicolumn{4}{c}{Layer 1} && \multicolumn{4}{c}{Layer 2} && \multicolumn{4}{c}{Layer 3} \\
    \cmidrule{2-16}
    & $\tau=$ & 1px$\downarrow$  & 3px$\downarrow$  & 5px$\downarrow$  & $\infty$px$\downarrow$ 
    && 1px$\downarrow$  & 3px$\downarrow$  & 5px$\downarrow$  & $\infty$px$\downarrow$  
    && 1px$\downarrow$  & 3px$\downarrow$  & 5px$\downarrow$  & $\infty$px$\downarrow$  \\
    \midrule
    RAFT \cite{raft} && 78.45 & 55.64 & 45.78 & \textbf{0.0}
                     && 100.00 & 100.00 & 100.00 & 100.00 
                     && 100.00 & 100.00 & 100.00 & 100.00 \\

    Multi-RAFT (L) && \textbf{76.51} & \textbf{51.82} & \textbf{42.63} & 9.19 
                   && 91.91 & 79.93 & 73.50 & 47.19
                   && 98.25 & 88.50 & 87.00 & 38.88 \\
    Multi-RAFT (S+L) && 77.83 & 54.85 & 45.39 & 8.83 
                     && \textbf{88.85} & \textbf{74.93} & \textbf{63.59} & \textbf{40.56} 
                     && \textbf{94.62} & \textbf{85.88} & \textbf{83.50} & \textbf{21.62} \\
    \bottomrule
  \end{tabular}
  }
  \caption{Multi-layer baseline and RAFT evaluated via multi-layer count-aware bad-$\tau$ on our benchmark, categorized by layer. Best results in \textbf{bold}.}
  \label{tab:evaluation_multilayer}
\end{table*}

\begin{table*}[h]
  \centering
  \resizebox{\linewidth}{!}{
  \begin{tabular}{l c ccc c ccc c ccc c ccc}
    \toprule
    \multirow{2}{*}{Method} && \multicolumn{3}{c}{All} && \multicolumn{3}{c}{Transparent} && \multicolumn{3}{c}{Reflective} && \multicolumn{3}{c}{Diffuse} \\
    \cmidrule{2-17}
    & $\tau=$ & 1px$\downarrow$  & 3px$\downarrow$  & 5px$\downarrow$  
    && 1px$\downarrow$  & 3px$\downarrow$  & 5px$\downarrow$  
    && 1px$\downarrow$  & 3px$\downarrow$  & 5px$\downarrow$  
    && 1px$\downarrow$  & 3px$\downarrow$  & 5px$\downarrow$ \\
    \midrule
    FlowNet-C \cite{flownet} && 94.94 & 77.37 & 64.39 
                             && 94.96 & 77.87 & 65.10 
                             && 94.89 & 78.26 & 65.62 
                             && 94.82 & 68.83 & 52.36 \\
    FlowNet2 \cite{flownet2} && 86.09 & 65.87 & 55.62 
                             && 87.45 & 67.69 & 57.45 
                             && 84.79 & 63.41 & 52.45 
                             && 71.70 & 48.18 & 39.56 \\
    PWC-Net \cite{pwc} && 83.77 & 63.09 & 53.42  
                       && 86.68 & 66.91 & 56.89 
                       && 74.06 & 50.42 & 42.20
                       && 69.15 & 43.79 & 35.11  \\
    GMA \cite{gma} && 78.93 & 56.27 & 45.62 
                   && 83.35 & 61.55 & 49.92  
                   && 65.12 & 40.63 & 32.77 
                   && 54.67 & 25.11 & 20.56  \\
    SKFlow \cite{skflow} && 78.64 & 56.57 & 47.36 
                         && 83.55 & 62.25 & 52.06 
                         && 62.37 & 38.13 & 31.93  
                         && 53.86 & 26.92 & 23.10  \\
    CRAFT \cite{craft} && 79.54 & 56.90 & 46.90  
                       && 84.00 & 62.17 & 51.23
                       && 64.59 & 40.21 & 33.22 
                       && 57.36 & 28.32 & 23.36  \\
    GMFlow \cite{gmflow} && 89.52 & 66.36 & 53.60
                         && 90.41 & 68.64 & 56.48 
                         && 88.76 & 62.08 & 45.13 
                         && 79.89 & 47.09 & 36.40  \\
    GMFlow+ \cite{gmflow+} && 89.52 & 66.36 & 53.60  
                           && 90.41 & 68.64 & 56.48  
                           && 88.76 & 62.08 & 45.13  
                           && 79.89 & 47.09 & 36.40 \\
    FlowFormer \cite{flowformer} && 78.20 & 57.54 & 47.94 
                                 && 83.03 & 63.31 & 53.23 
                                 && 62.17 & 39.66 & 31.50 
                                 && 53.74 & 25.40 & \textbf{18.57} \\
    RAFT \cite{raft} && 78.13 & 54.99 & 44.85 
                     && 83.20 & 60.21 & 49.02
                     && \underline{61.68} & 39.38 & 32.18
                     && \underline{51.66} & \textbf{24.57} & \underline{21.06} \\
    \midrule
    Multi-RAFT (L)   && \textbf{75.37} & \textbf{48.98} & \textbf{38.73} 
                     && \textbf{79.35} & \textbf{52.78} & \textbf{41.32} 
                     && 61.91 & \underline{36.86} & \underline{30.85}
                     && 55.89 & 28.75 & 23.89 \\
    Multi-RAFT (S+L) && \underline{76.38} & \underline{51.64} & \underline{40.95}  
                     && \underline{81.41} & \underline{56.45} & \underline{44.43} 
                     && \textbf{60.44} & \textbf{36.76} & \textbf{30.46} 
                     && \textbf{49.22} & \underline{24.83} & \underline{21.01} \\
    \bottomrule
  \end{tabular}}
  \caption{Multi-layer baseline and other methods evaluated via multi-layer  bad-$\tau$ on our benchmark, categorized by material. Best results in \textbf{bold}, second best \underline{underlined}.}
  \label{tab:evaluation_multilayer_all}
\end{table*}

\begin{figure*}[h]
    \centering
    \includegraphics[width=0.8\linewidth]{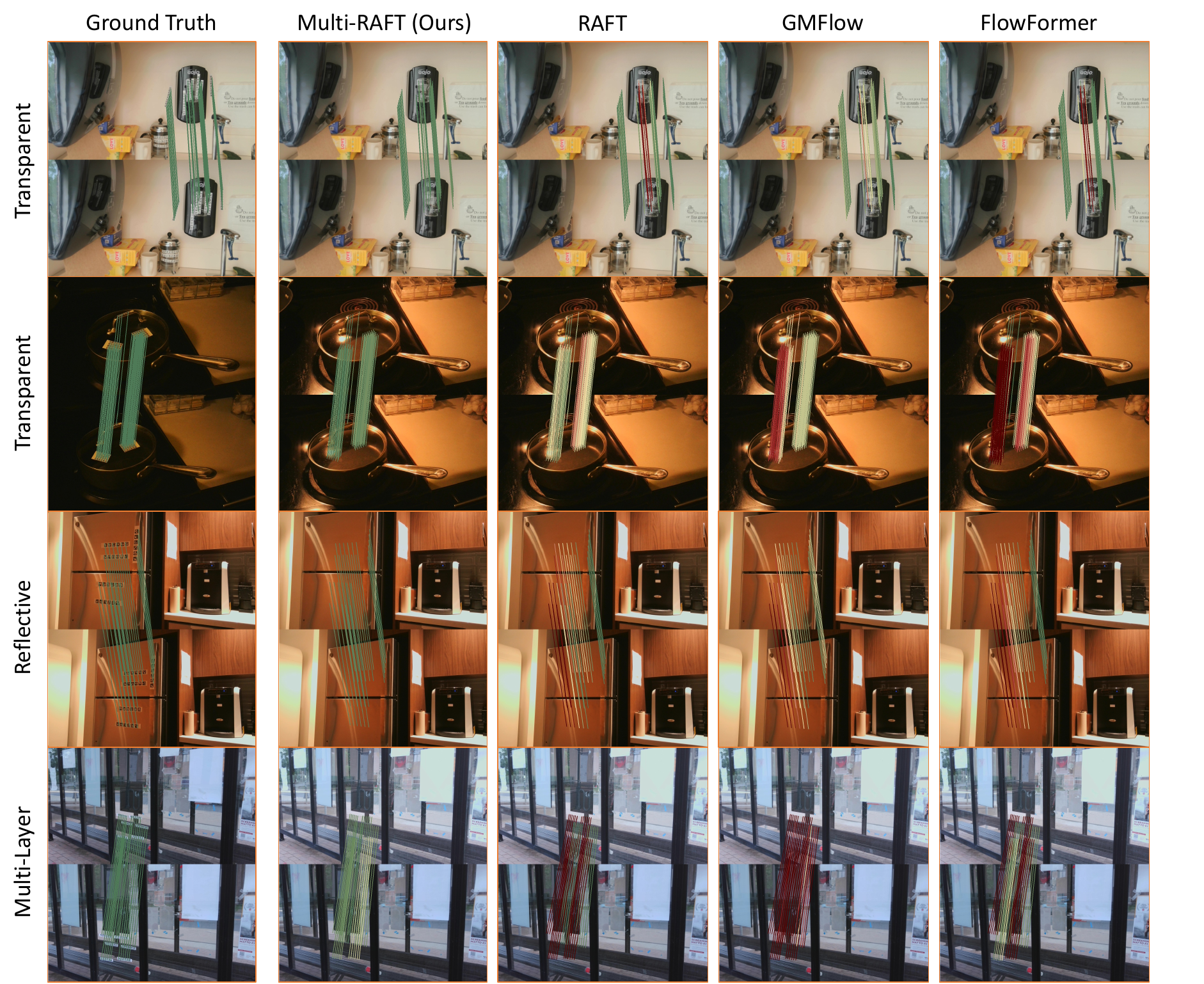}
    \caption{Qualitative results of our baseline model and three representative optical flow methods. Lines in images show flow predictions, with greener lines indicating smaller L2 error and redder lines indicating larger L2 error. The first two rows, third row and fourth row show results in transparent, reflective, and multi-layer settings.}
    \label{fig:baseline_visualization}
\end{figure*}

We also compare our baseline model to existing single-layer optical flow methods directly using multi-layer bad-$\tau$. By default, these existing methods are unable to predict multiple layers, incurring 100\% error rate for layers beyond the first. To enable fairer comparisons, we apply a workaround for existing optical flow methods. Specifically for each pixel, we take the model's single flow vector prediction and pretend that the model produced that flow prediction for each layer that has ground truth optical flow annotation. This adjustment helps us perform evaluation with existing methods on every annotated pixel in our benchmark, regardless of which layer it is on. Because we disregard layer count accuracy, we disable the flow pruning of our baseline method. 

Results are shown in \cref{tab:evaluation_multilayer_all}. Even in this unfair setting, our baseline outperforms all existing optical flow methods on non-Lambertian surfaces by a large margin and maintains comparable performance on diffuse surfaces. Qualitative results are shown in \cref{fig:baseline_visualization}. Our baseline model has better understanding of non-Lambertian surfaces; for example, only our Multi-RAFT model accurately predicts optical flow on both transparent materials and opaque regions. However, our baseline still has much room for improvement, and we hope \projectname{} will encourage further research on non-Lambertian optical flow. 

\section{Conclusion}
We proposed \projectname{}, a diverse multi-layer real world benchmark for non-Lambertian optical flow, a large-scale synthetic dataset and a baseline. We believe our data will fuel the field of 3D non-Lambertian object understanding.
\label{sec:conclusion}

\newpage
\section*{Acknowledgements}

This work was partially supported by the National Science Foundation. Erich Liang was supported by NSF GRFP (2146752).  
We additionally thank support from and helpful discussions with friends and colleagues at Princeton University.
We extend our special thanks to Professor Yibin Kang and his students for their assistance with data collection in the molecular biology laboratory.

\newpage
\section{Appendix}

\subsection{Details about \projectname{} Benchmark Data Collection}

\subsubsection{Data Statistics}

Detailed statistics are shown in \cref{tab:stats_benchmark}.

\begin{table*}[b]
  \centering
    \begin{tabular}{lccccccc}
   \toprule
   & \multirow{2}{*}{All} & \multicolumn{3}{c@{}}{Material}& \multicolumn{3}{c@{}}{Layer} \\
   \cmidrule(l){3-8}  
   & & \mc{Transparent} \ & \mc{Reflective} \ & \mc{Diffuse} \ & \mc{1} & \mc{2} & \mc{3} \\
   \midrule 

   Optical Flow \ & 152627 & 120737 & 21459 & 10431 & 136799 & 13988 & 1840 \\
   Stereo & 147607 & 117262 & 20479 & 9866 & 132991 & 13048 & 1568 \\
   \bottomrule
\end{tabular}
\caption{Number of stereo pairs and optical flow pairs in our benchmark, categorized by material property and layer index.}
\label{tab:stats_benchmark}
\end{table*}

\subsubsection{Ground Truth Annotation}
We used the Python bindings of AprilTag \cite{AprilTag-python} to detect of AprilTag \cite{apriltags} in raw images.

Upon detection, each AprilTag marker is uniquely identified by its ID, the central point of the tag, and the locations of its four corners. To reduce potential errors arising from distortion, especially because some markers are attached to curved surfaces, we limit the generation of ground-truth correspondences to only the four corners of each tag. Consequently, each tag offers four pairs of correspondences.

\subsubsection{Camera Calibration}
The image acquisition process involves two cameras mounted on a tripod. Prior to each image capture session, the cameras are calibrated. For each camera calibration session, we take at least 40 pairs of simultaneous photos with the camera pair. We utilize OpenCV\cite{opencv} library to interface with each camera.

For each camera's image of the chessboard, we identify the 2D key points --- the inner corners of the chessboard. By measuring the size of each square in the chessboard, we are able to determine the position of each corner in world coordinates. The relationship between 2D and 3D key points in homogeneous coordinates is represented by the equation:
\begin{align}
    \mathbf{P}_{2D} = \mathbf{K} [\mathbf{R} \mid \mathbf{t}] \mathbf{P}_{3D}
\end{align}
Here, $\mathbf{P}_{2D}$ and $\mathbf{P}_{3D}$ denote the 2D and 3D key points, respectively. $\mathbf{K}$ is the camera's intrinsic matrix, and $[\mathbf{R} \mid \mathbf{t}]$ is the camera's extrinsic matrix. This allows us to solve the camera's intrinsic matrices $\mathbf{K}$ and distortion coefficients.

Subsequently, the two cameras undergo stereo calibration. We assume that the origin of world coordinates is located at the center of the left camera, and we calculate $\mathbf{R}_{cam}$ and $\mathbf{t}_{cam}$, which represent the rotation and translation from the left camera to the right camera. This is done by jointly calibrating $\mathbf{P}_{3D}$, $\mathbf{P}_{2D_l}$, and $\mathbf{P}_{2D_r}$, where $\mathbf{P}_{2D_l}$ and $\mathbf{P}_{2D_r}$ are the 2D key points on the left and right images, respectively. Utilizing all this information, we proceed to rectify the images, ensuring the corresponding points in the two images lie along the same epipolar lines.

\subsection{Details about Synthetic Data Generation}

\subsubsection{Scenes and Assets}
Our synthetic dataset was created using 30 diverse scenes, enhanced with 100 non-Lambertian assets and 50 random HDR environment textures. 
All scenes and assets were acquired from BlendSwap \cite{BlendSwap} under the Creative Commons license. Note that some of the assets are adopted from the other scenes. We acknowledge creators of all assets and scenes, shown in \cref{tab:asset}. All HDR images are acquired from HDRi Haven \cite{HDRI-haven} under the Creative Commons Zero license.

\begin{table}[]
    \centering
    \resizebox{\linewidth}{!}{
    \begin{tabular}{ccccccccc}
       Type & Category & Link & Creator && Type & Category & Link & Creator\\
       \toprule
        Scene & Kitchen & \href{https://blendswap.com/blend/21884}{link}& TheCGNinja  
        && Scene & Living Room & \href{https://www.blendswap.com/blend/13491}{link} &  Wig42 \\
        Scene & Kitchen & \href{https://blendswap.com/blend/7489}{link}& cenobi 
        && Scene & Living Room & \href{https://blendswap.com/blend/8683}{link} & Mikel007 \\
        Scene & Kitchen & \href{https://blendswap.com/blend/22739}{link} & Warcos 
        && Scene & Living Room & \href{https://blendswap.com/blend/11811}{link} &  blenderjunky \\
        Scene & Kitchen & \href{https://blendswap.com/blend/4722}{link} & unangelo
        && Scene & Living Room & \href{https://blendswap.com/blend/6568}{link} & ermmus \\
        Scene & Kitchen & \href{https://blendswap.com/blend/10416}{link} & oldtimer 
        && Scene & Living Room & \href{https://blendswap.com/blend/8381}{link} & oldtimer \\
        Scene & Kitchen & \href{https://blendswap.com/blend/11801}{link} & blenderjunky 
        && Scene & Bedroom & \href{https://blendswap.com/blend/3391}{link} & SlykDrako \\
        Scene & Kitchen & \href{https://blendswap.com/blend/17979}{link} & MarcoD 
        && Scene & Bedroom & \href{https://blendswap.com/blend/5777}{link} & irokrhus \\
        Scene & Kitchen & \href{https://blendswap.com/blend/5472}{link} & MimingApe 
        && Scene & Bedroom & \href{https://blendswap.com/blend/12608}{link} & oldtimer \\
        Scene & Kitchen & \href{https://blendswap.com/blend/7914}{link} & oldtimer 
        && Scene & Bedroom & \href{https://blendswap.com/blend/6501}{link} & Yulia \\
        Scene & Kitchen & \href{https://blendswap.com/blend/8366}{link} & appisolato 
        && Scene & Bedroom & \href{https://blendswap.com/blend/26193}{link} & Mikel007 \\
        Scene & Office & \href{https://blendswap.com/blend/19984}{link} & ThePefDispenser
        && Assets & N/A & \href{https://www.blendswap.com/blend/20700}{link} & ruwo \\
        Scene & Office & \href{https://blendswap.com/blend/18784}{link} & LRosario 
        && Assets & N/A & \href{https://www.blendswap.com/blend/26016}{link} & Davilion \\
        Scene & Office & \href{https://www.blendswap.com/blend/17849}{link} & DragonautX 
        && Assets & N/A & \href{https://www.blendswap.com/blend/18423}{link} & MZiemys \\        
        Scene & Office & \href{https://www.blendswap.com/blend/16529}{link} & fjcar 
        && Assets & N/A & \href{https://www.blendswap.com/blend/18423}{link} & MZiemys \\
        Scene & Office & \href{https://blendswap.com/blend/19547}{link} & Elysia 
        && Assets & N/A & \href{https://www.blendswap.com/blend/26016}{link} & Davilion \\
        Scene & Bathroom & \href{https://blendswap.com/blend/3865}{link} &  bobal57 
        && Assets & N/A & \href{https://www.blendswap.com/blend/18935}{link} & Zorian \\
        Scene & Bathroom & \href{https://blendswap.com/blend/5755}{link} & irokrhus 
        && Assets & N/A & \href{https://www.blendswap.com/blend/3893}{link} & vicentecarro \\
        Scene & Bathroom & \href{https://blendswap.com/blend/8180}{link} & wfg5001 
        && Assets & N/A & \href{https://www.blendswap.com/blend/17550}{link} & piergi \\
        Scene & Bathroom & \href{https://blendswap.com/blend/12584}{link} & nacimus 
        && Assets & N/A & \href{https://www.blendswap.com/blend/12590}{link} & Bastable \\
        Scene & Bathroom & \href{https://blendswap.com/blend/18595}{link} & Ndakasha 
        &&  Assets & N/A & \href{https://www.blendswap.com/blend/11336}{link} & arttechsouth \\
    \end{tabular}}
    \caption{Blender assets and scenes.}
    \label{tab:asset}
\end{table}

\subsubsection{Ground Truth Generation Details}
To generate ground truth for optical flow, a typical approach involves using the vector pass in the Blender Cycles engine\cite{blender}. However, Cycles does not inherently support the generation of multi-layer ground truth. To address this limitation, we add several new passes to the engine, enabling it to record information each time a ray strikes a surface during the ray tracing process. Specifically, we modified the Cycles source code to capture data for multiple layer masks, 3D positions (useful for depth and disparity calculations), and motion (for optical flow calculation) each time a ray from air strikes an object surface. This modification allows for the generation of multi-layer ground truth that is perfectly aligned with human perception and preserves the effects of light refraction.

\subsection{Training Details}
All models are implemented in PyTorch \cite{PyTorch}. The fine-tuned version of RAFT is trained on eight RTX 3090 GPUs with a batch size of 20, directly following the training procedure and data augmentation in RAFT \cite{raft}. The learning rate is set to 1e-5. 

For multi-layer RAFT is trained on four RTX 3090 GPUs with a batch size of 4. The learning rate is set to 1e-4. When the training images contain $m \ge 1$ layers of true optical flow and the model generates $n > m$ optical flow prediction layers, the final prediction layer is duplicated $n - m + 1$ times to align the dimensions. Specifically, for training with the Sintel \cite{sintel}, which provides a single layer of optical flow ground truth, this duplication occurs $n$ times.

For (S+L) training policy, any image in Sintel dataset will appear 100 times to match the size of our synthetic dataset. The reported results from the checkpoint that has the best performance on validation set of our benchmark.

\subsection{Additional Experiments}

\subsubsection{First Layer Optical Flow}

\begin{table*}[t]
  \centering
  \resizebox{\linewidth}{!}{
  \begin{tabular}{l c cccc c cccc c cccc c cccc}
    \toprule
    \multirow{2}{*}{Method} && \multicolumn{4}{c}{All} && \multicolumn{4}{c}{Transparent} && \multicolumn{4}{c}{Reflective} && \multicolumn{4}{c}{Diffuse} \\
    \cmidrule{2-21}
    && EPE$\downarrow$ & 1px$\downarrow$ & 3px$\downarrow$ & 5px$\downarrow$ 
    && EPE$\downarrow$ & 1px$\downarrow$ & 3px$\downarrow$ & 5px$\downarrow$ 
    && EPE$\downarrow$ & 1px$\downarrow$ & 3px$\downarrow$ & 5px$\downarrow$
    && EPE$\downarrow$ & 1px$\downarrow$ & 3px$\downarrow$ & 5px$\downarrow$\\
    \midrule
    FlowNet-C \cite{flownet} && 9.71 & 89.07 & 61.51 & 43.93 && 11.08 & 89.23 & 62.43 & 45.05 && 6.38 & 88.25 & 58.36 & 40.03 && 8.53 & 89.08 & 54.13 & 35.35 \\
    FlowNet2 \cite{flownet2} && 10.07 & 77.56 & 54.22 & 42.13 &&
                11.46 & 78.20 & 56.15 & 44.38 &&
                6.70 & 75.39 & 46.69 & 33.33 &&
                7.66 & 72.44 & 43.21 & 29.41 \\
    PWC-Net \cite{pwc} && 9.49 & 74.93 & 50.47 & 39.05
                && 10.90 & 76.47 & 52.59 & 41.43
                && 5.99 & 69.84 & 42.99 & 29.82
                && 6.91 & 61.85 & 34.77 & 25.30 \\
    GMA \cite{gma} && 9.77 & 72.46 & 46.93 & 36.97
            && 12.01 & 75.48 & 50.07 & 40.24
            && \textbf{4.48} & 60.85 & 35.42 & 24.26
            && 2.26 & 54.20 & 25.56 & 17.98 \\
    SKFlow \cite{skflow} && 9.86 & 72.02 & 47.44 & 36.88
            && 12.00 & 74.90 & 50.84 & 40.14
            && 4.78 & 60.89 & 35.21 & 24.40
            && 3.23 & 54.90 & 23.23 & 17.18 \\
    CRAFT \cite{craft} && 10.36 & 72.34 & 47.54 & 37.00
            && 12.65 & 74.75 & 50.96 & 40.47
            && 4.65 & 64.12 & 35.10 & 23.06
            && 3.30 & 53.08 & 23.95 & 18.89\\
    GMFlow \cite{gmflow} && 9.09 & 81.99 & 51.79 & 37.75
            && 10.93 & 83.01 & 53.87 & 40.06
            && 5.20 & 80.02 & 44.73 & 28.64
            && 5.01 & 66.91 & 35.13 & 24.79 \\
    GMFlow+ \cite{gmflow+} && 9.46 & 82.71 & 53.14 & 39.70
            && 11.31 & 83.21 & 54.91 & 42.10
            && 6.04 & 81.61 & 46.57 & 29.95
            && 5.71 & 75.97 & 41.43 & 27.85\\
    FlowFormer \cite{flowformer} && 10.20 & 73.59 & 48.97 & 38.56
            && 12.51 & 76.91 & 52.56 & 42.27
            && 5.00 & 61.03 & 36.18 & 24.76
            && \textbf{2.17} & 52.89 & 22.90 & 14.12 \\ 
    RAFT \cite{raft} && 9.38 & 71.98 & 46.46 & 36.15
            && 11.31 & 74.65 & 49.69 & 39.34
            && 5.57 & 61.53 & 35.73 & 24.57
            && 2.62 & 56.72 & \textbf{19.44} & \textbf{14.05} \\
    \midrule
    RAFT-ft. (S) && 9.74 & 74.56 & 49.14 & 38.60
            && 11.64 & 77.10 & 51.94 & 41.55
            && 5.74 & 65.16 & 39.28 & 27.47
            && 4.13 & 57.63 & 28.32 & 20.06 \\
    RAFT-ft. (L) && \underline{\textbf{7.12}} & \underline{\textbf{69.17}} & \underline{\textbf{40.88}} & \underline{\textbf{29.49}}
            && \underline{\textbf{8.26}} & \underline{\textbf{71.74}} & \underline{\textbf{43.44}} & \underline{\textbf{31.79}}
            && \underline{5.24} & \underline{59.72} & \underline{\textbf{31.23}} & \underline{\textbf{20.60}}
            && 3.03 & \underline{\textbf{51.77}} & 24.61 & 15.84 \\
    RAFT-ft. (S+L) && \underline{7.93} & \underline{69.20} & \underline{42.04} & \underline{32.51}
            && \underline{9.23} & \underline{71.88} & \underline{44.76} & \underline{35.05}
            && 6.16 & \underline{\textbf{58.68}} & \underline{32.42} & \underline{22.55}
            && 2.65 & \underline{54.13} & 22.02 & 18.27 \\
    \bottomrule
  \end{tabular}
  }
  \vspace{-3mm}
  \caption{Representative optical flow methods evaluated on first layer subset of our benchmark using EPE and bad-$\tau$ metrics. Images are down-sampled by 8. Best scores are in \textbf{bold}. \underline{Underlined numbers} denote RAFT fine-tuned on our synthetic data outperforming the original version.}
  \label{tab:optical_flow_first_layer}
\end{table*}

We provide additional evaluation results for our single layer experiments. For this set of experiments, we evaluate each method to predict first layer optical flow on pairs of images that have been downsampled by a factor of 8---this is as opposed to our results in the main paper, where we evaluate each method on images downsampled by a factor of 4. Overall, our finetuned RAFT method still outperforms other existing optical flow methods, including the baseline RAFT method. Results are shown in Tab. \ref{tab:optical_flow_first_layer}.

\subsubsection{First-Layer Stereo Matching} 

As mentioned in main paper, our benchmark also provides stereo matching ground-truth.
We evaluate effectiveness of existing representative stereo matching methods with public implementation and pre-trained weights on \projectname{}'s first layer points. Stereo pairs with significant $y$-axis discrepancies are excluded, achieving an average residual $y$-disparity of 0.36 on images downsampled by a factor of 4 to $540 \times 960$. Results are shown in \cref{tab:stereo_first_layer}. 

As stereo matching methods tend to be sensitive to the scale of images, we also provide results on images that have been downsampled by a factor of 8, shown in \cref{tab:stereo_first_layer_8px}. Overall, existing methods generally struggle to achieve good EPE and bad-$\tau$ metrics, particularly for transparent and reflective materials. This highlights the challenge of first-layer stereo matching in non-Lambertian settings.

\begin{table*}[h]
  \centering
  \resizebox{\linewidth}{!}{
  \begin{tabular}{l c cccc c cccc c cccc c cccc}
    \toprule
    \multirow{2}{*}{Method} && \multicolumn{4}{c}{All} && \multicolumn{4}{c}{Transparent} && \multicolumn{4}{c}{Reflective} && \multicolumn{4}{c}{Diffuse} \\
    \cmidrule{2-21}
    && EPE$\downarrow$ & 1px$\downarrow$ & 3px$\downarrow$ & 5px$\downarrow$ 
    && EPE$\downarrow$ & 1px$\downarrow$ & 3px$\downarrow$ & 5px$\downarrow$ 
    && EPE$\downarrow$ & 1px$\downarrow$ & 3px$\downarrow$ & 5px$\downarrow$
    && EPE$\downarrow$ & 1px$\downarrow$ & 3px$\downarrow$ & 5px$\downarrow$\\
    \midrule
    PSMNet \cite{psmnet} && 74.43 & 92.16 & 83.32 & 77.78 
            && 82.92 & 95.57 & 89.68 & 84.78 
            && 45.41 & 82.62 & 62.38 & 53.70 
            && 17.82 & 59.91 & 37.45 & 31.62 \\
    HSMNet \cite{hsmnet} && 57.38 & 93.91 & 87.43 & 83.03 
            && 64.48 & 98.70 & 95.19 & 91.71 
            && 33.74 & 81.14 & 63.72 & 54.86 
            && 7.40 & 45.96 & 23.08 & 18.49 \\
    LEAStereo \cite{leastereo} && 54.96 & 89.58 & 79.20 & 74.36 
            && 62.19 & 95.66 & 87.51 & 82.96 
            && 29.56 & 70.47 & 50.54 & 43.96 
            && 9.66 & 41.80 & 24.86 & 21.45 \\
    CFNet \cite{cfnet} && 40.23 & 90.45 & 83.09 & 77.79 
            && 45.90 & 95.40 & 90.27 & 85.61 
            && 20.07 & 75.16 & 59.53 & 51.75 
            && 5.68 & 50.27 & 30.85 & 22.49 \\
    PCWNet \cite{pcwnet} && 41.59 & 92.00 & 84.23 & 79.58 
            && 47.44 & 97.55 & 92.81 & 88.37 
            && 20.90 & 75.80 & 55.93 & 50.01 
            && 5.68 & 42.77 & 22.53 & 18.92 \\
    RAFTStereo \cite{raft-stereo}&& 32.50 & 85.27 & 75.80 & 71.25 
            && 37.36 & 92.99 & 85.04 & 80.22 
            && 13.94 & 60.15 & 42.45 & 38.61 
            && 8.55 & 28.01 & 21.83 & 20.01 \\
    DLNR \cite{DLNR} && \textbf{30.69} & \textbf{82.47} & \textbf{71.92} & \textbf{67.24} 
            && \textbf{36.10} & \textbf{90.48} & \textbf{81.64} & \textbf{76.57} 
            && \textbf{9.96} & \textbf{56.44} & \textbf{37.02} & \textbf{33.87} 
            && \textbf{4.49} & \textbf{23.12} & \textbf{14.49} & \textbf{11.58} \\
    \bottomrule
  \end{tabular}
  }
  \vspace{-3mm}
  \caption{Representative stereo matching methods evaluated on first layer subset of our benchmark using EPE and bad-$\tau$ metrics. Best scores are in \textbf{bold}.}

  \label{tab:stereo_first_layer}
\end{table*}

\begin{table*}[h]
  \centering
  \resizebox{\linewidth}{!}{
  \begin{tabular}{l c cccc c cccc c cccc c cccc}
    \toprule
    \multirow{2}{*}{Method} && \multicolumn{4}{c}{All} && \multicolumn{4}{c}{Transparent} && \multicolumn{4}{c}{Reflective} && \multicolumn{4}{c}{Diffuse} \\
    \cmidrule{2-21}
    && EPE$\downarrow$ & 1px$\downarrow$ & 3px$\downarrow$ & 5px$\downarrow$ 
    && EPE$\downarrow$ & 1px$\downarrow$ & 3px$\downarrow$ & 5px$\downarrow$ 
    && EPE$\downarrow$ & 1px$\downarrow$ & 3px$\downarrow$ & 5px$\downarrow$
    && EPE$\downarrow$ & 1px$\downarrow$ & 3px$\downarrow$ & 5px$\downarrow$\\
    \midrule
    PSMNet \cite{psmnet} && \textbf{14.73} & 84.77 & \textbf{63.11} & \textbf{51.82}
            && \textbf{16.44} & \textbf{88.89} & \textbf{69.15} & \textbf{57.37} 
            && 8.11 & 72.52 & 41.74 & 31.42 
            && 4.77 & 44.55 & 19.38 & 15.16  \\
    HSMNet \cite{hsmnet}&& 18.70 & 89.83 & 76.93 & 66.72 
            && 20.83 & 94.79 & 84.36 & 74.55 
            && 11.25 & 75.62 & 51.69 & 37.92 
            && 3.04 & 39.43 & 18.45 & 14.75 \\
    LEAStereo \cite{leastereo}&& 15.51 & 82.92 & 64.61 & 54.40 
            && 17.77 & 89.44 & 71.92 & 61.46 
            && 6.71 & 61.42 & 38.07 & 27.04 
            && 2.81 & 28.83 & 14.64 & 13.59 \\
    CFNet \cite{cfnet} && 16.38 & 84.81 & 70.06 & 60.20 
            && 18.39 & 91.07 & 76.72 & 66.67 
            && 9.26 & 65.46 & 47.98 & 37.85 
            && 1.98 & 27.18 & 15.38 & 10.75 \\
    PCWNet \cite{pcwnet} && 17.56 & 89.44 & 76.35 & 66.52 
            && 19.86 & 96.20 & 84.27 & 74.42 
            && 9.14 & 69.14 & 49.40 & 38.17 
            && 2.21 & 24.57 & 14.38 & 11.13 \\
    RAFTStereo \cite{raft-stereo} && 16.72 & 84.14 & 70.34 & 59.71 
            && 19.05 & 91.48 & 78.62 & 67.21 
            && 7.75 & 59.88 & 40.47 & 31.44 
            && 2.82 & 23.45 & 13.03 & 12.99  \\
    DLNR \cite{DLNR} && 15.91 & \textbf{82.06} & 68.79 & 60.69 
            && 18.70 & 91.15 & 78.28 & 70.24 
            && \textbf{4.80} & \textbf{49.88} & \textbf{33.58} & \textbf{24.25} 
            && \textbf{0.92} & \textbf{16.50} & \textbf{7.21} & \textbf{3.10} \\

    \bottomrule
  \end{tabular}
  }
  \vspace{-3mm}
  \caption{Representative stereo matching methods evaluated on first layer subset of our benchmark using EPE and bad-$\tau$ metrics. Images are down-sampled by 8. Best scores are in \textbf{bold}.}

  \label{tab:stereo_first_layer_8px}
\end{table*}

\bibliographystyle{splncs04}
\bibliography{egbib}
\end{document}